\pdfoutput=1

\documentclass[11pt]{article}

\usepackage[final]{acl}

\usepackage{times}
\usepackage{latexsym}

\usepackage[T1]{fontenc}

\usepackage[utf8]{inputenc}

\usepackage{microtype}

\usepackage{inconsolata}

\usepackage{graphicx}

\usepackage{booktabs}
\usepackage{multirow}
\usepackage{amsmath}
\usepackage{amssymb}
\usepackage{amsfonts}
\usepackage{bbm}
\usepackage{diagbox}
\usepackage{graphicx}
\usepackage{makecell}
\usepackage{xspace}
\usepackage{mathabx}
\usepackage[normalem]{ulem}
\usepackage{caption,subcaption}
\usepackage{stackengine}

\usepackage{hyperref}       
\usepackage{url}            
\usepackage[dvipsnames]{xcolor}      

\usepackage{nicefrac}       

\usepackage{lineno}

\usepackage{wrapfig}
\usepackage{makecell}

\definecolor{darkblue}{rgb}{0, 0, 0.5}
\hypersetup{colorlinks=true, citecolor=darkblue, linkcolor=darkblue, urlcolor=darkblue}

\newcommand{\base}{\textsl{base form}\xspace}
\newcommand{\bases}{\textsl{base forms}\xspace}
\newcommand{\transformation}{\textsl{transformation}\xspace}
\newcommand{\transformations}{\textsl{transformations}\xspace}

\newcommand{\patchscopes}{{Patchscopes}\xspace}

\usepackage[normalem]{ulem}

\newcommand{\resolved}[1]{}

\newcommand{\tabref}[1]{Table~\ref{tab:#1}}
\newcommand{\figref}[1]{Figure~\ref{fig:#1}}
\newcommand{\appref}[1]{Appendix~\ref{#1}}
\newcommand{\secref}[1]{§\ref{sec:#1}}

\newcommand{\equref}[1]{Eq.~\ref{eq:#1}}

\title{
Vocab Diet: Reshaping the Vocabulary of LLMs via Vector Arithmetic
}

\author{Yuval Reif \;\;\;\;\; Guy Kaplan \;\;\;\;\; Roy Schwartz\\
  The Hebrew University of Jerusalem \\
  \texttt{\normalsize \{yuval.reif,guy.kaplan3,roy.schwartz1\}@mail.huji.ac.il}}

\begin{document}
\maketitle

\defcitealias{gemma}{Gemma Team, 2024}
\defcitealias{olmo2}{OLMo Team, 2024}

\begin{abstract}

Large language models~(LLMs) often encode word-form variation (e.g., \textit{walk} vs.\ \textit{walk\textbf{ed}}) as linear directions in the embedding space. However, standard tokenization algorithms treat such variants as distinct words with different vocabulary entries---quickly filling the size-capped token vocabulary with surface-form variation~(e.g., \textit{walk}, \textit{walk\textbf{ing}}, \textit{\textbf{W}alk}), at the expense of diversity and multilingual coverage.
We show that many of these variations can be captured by \textsl{\transformation vectors}---additive offsets that yield the appropriate word representation when applied to a \base embedding, in both the input and output spaces.
Building on this, we propose a compact reshaping of the vocabulary: instead of assigning unique tokens to each surface form, we compose them from shared \base and \transformation vectors~(e.g., \textit{walked} is \textit{walk}$+$\textsl{past tense}). Our approach is lightweight---keeping the pretrained backbone frozen and only training small adaptation modules.
We apply it across five languages and multiple LLMs in both pretraining and post-hoc adaptation, freeing 10--40\% of vocabulary slots to be reallocated where tokenization is inefficient.
Importantly, we do so while also expanding vocabulary coverage to out-of-vocabulary words, and with minimal impact on downstream performance.
Our findings motivate a rethinking of vocabulary design, towards a representation that better matches the underlying structure of language and the practical needs of multilingual coverage.\footnote{Code is available at \url{https://vocabdiet.github.io}.}

\end{abstract}

\section{Introduction}

\begin{figure}[!ht]
   \centering
   \includegraphics[width=0.82\columnwidth]{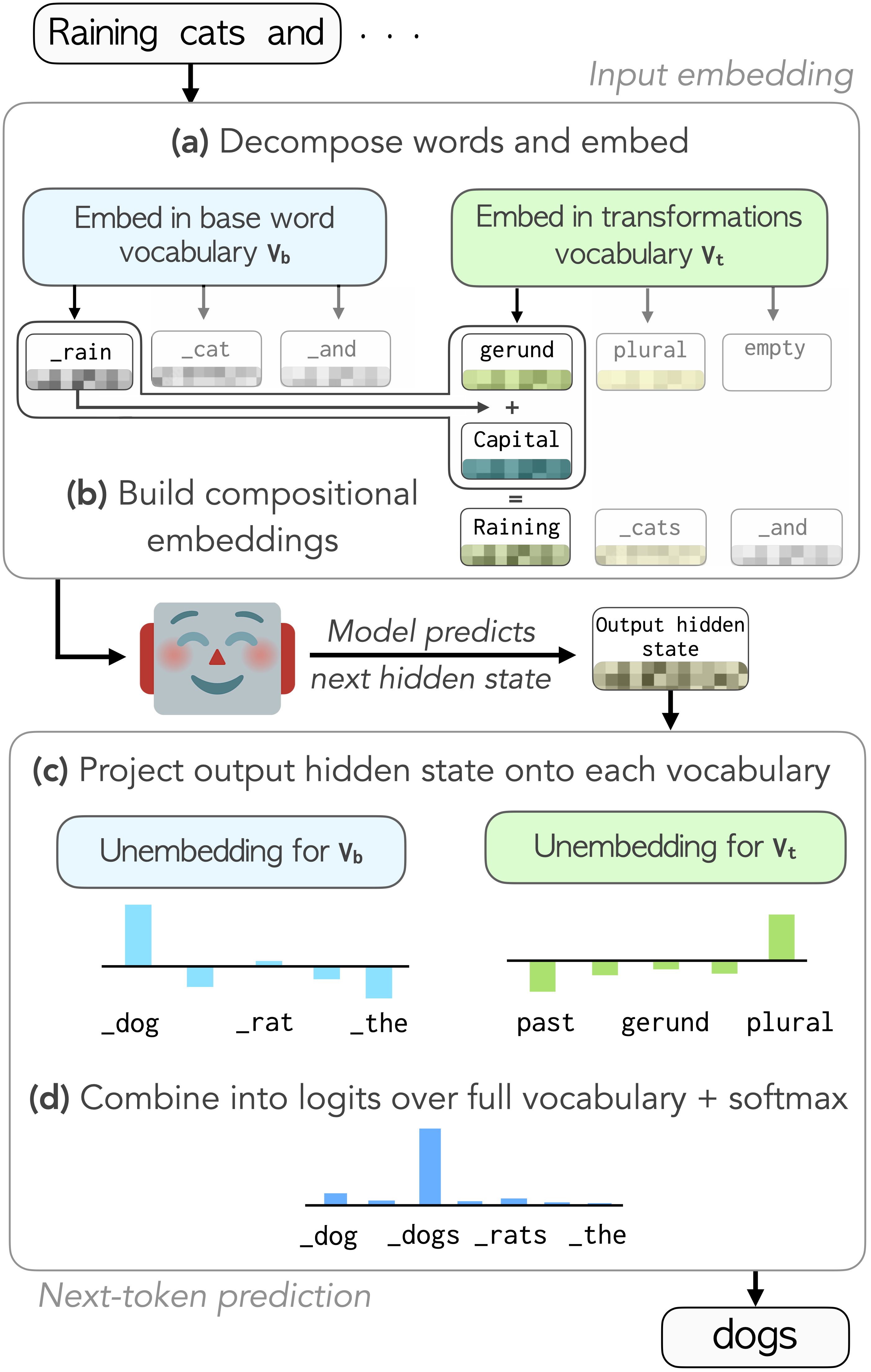}\caption{
\textbf{Compositional vocabulary for LLMs.}  
\textbf{Top:} Input tokens are represented by \textbf{(a)} decomposing them into base words (\textcolor{Cerulean}{$\mathcal{V}_b$}) and transformations (\textcolor{LimeGreen}{$\mathcal{V}_t$}), and \textbf{(b)} feeding the composite embeddings to the model. For example, ``\textit{cats}'' becomes \textit{cat} + \textsl{plural}. \textbf{Bottom:} The next token is predicted by \textbf{(c)} computing logits independently over base words and transformations, and \textbf{(d)} combining them into next-token probabilities. 
Our approach works seamlessly both as a lightweight adaptation of pretrained LLMs and when pretraining from scratch, creating a more compact vocabulary that supports a wider array of words.
}
   \label{fig:illustration}
\end{figure}

Modern large language models (LLMs) typically rely on subword tokenization algorithms like byte-pair encoding~\citep[BPE;][]{bpe2016sennrich}. Such methods allocate tokens to frequent words and split less~frequent ones into sequences of sub-word tokens---minimizing the number of tokens needed to represent typical textual data. 
Recent models use ever-larger vocabularies, often exceeding 100k tokens~\citep{llama3,openai2024tiktoken,qwen2.5}. While recent work calls for scaling up the vocabulary even further~\citep{tao2024scaling, huang2025vocabscaling}, the computational cost of supporting large vocabularies forces developers to cap its size~\citep{dagan2024getting,wijmans2025cut}.
Vocabulary design is therefore a resource allocation problem: every slot added to one language or domain comes at the expense of coverage and efficiency elsewhere~\citep{foroutan-meister-et-al-2025-parity-aware-bpe}.

Standard tokenization, while effective, often leads to a disproportionate allocation of the vocabulary~(\secref{vocabulary_redundancy}). Common words occupy multiple token slots for their various forms (e.g., \textit{walk}, \textit{walks}, \textit{walking}), leaving less room for uncommon words and multilingual coverage. This approach ultimately hurts both performance and inference costs~\citep{petrov2023language, ahia2023languages,tokenizer-choice-2024}. More fundamentally, it ignores a striking property of LLMs: their tendency to encode relationships between words as simple \emph{linear directions}~\citep{park2024linearrepresentationhypothesisgeometry,marks2024truth}.
Our central question is whether this structure can be leveraged to build more compact and expressive vocabularies under a fixed size---allowing for more efficient tokenization across domains.
 
We begin by investigating how LLMs represent word form variation. Building on the idea of vector arithmetic in embedding space~\citep{mikolov-etal-2013-linguistic-regularities}, we examine whether common word-form transformations—including morphological inflection (\textit{walk\textbf{ed}}), derivation (\textit{walk\textbf{able}}) and capitalization (\textit{\textbf{W}alk})—can be captured as consistent \textsl{\transformation vectors} added to a \textsl{\base} word embedding~(\secref{method}). 
Focusing on five morphologically diverse languages, we identify token pairs of base- and surface-form words exemplifying the same relation using UniMorph~\citep{batsuren-etal-2022-unimorph}. We then compute the average offset vector for each relation, and use these as \transformation vectors. 
Our results show that adding these vectors to \base embeddings yields representations that the model interprets similarly to the expected surface form~\citep{ghandeharioun2024patchscopes}. 
Interestingly, this holds even when the target word is not represented as a single token in the vocabulary,\footnote{E.g., a word like ``walkable'' is split into \texttt{[\_walk, able]}.} indicating that LLMs process and interpret word forms compositionally~(\secref{does_it_work}).

Building on these insights, we propose a compact restructuring of the vocabulary, building word embeddings from shared components~(\figref{illustration}): a \textsl{\base} vector for the core lexical item and a \textsl{\transformation} vector for encoding word-form variation. Rather than assigning a unique token embedding to each surface form, we remove inflected forms from the model’s embedding tables. Instead, we introduce a small set of \transformation embeddings---enabling us to represent the discarded words compositionally~(e.g., \textit{walked} as \textsl{walk}$+$\textsl{past tense}) in both input and output. 

We study two practical regimes: lightweight post-hoc fine-tuning and compositional pretraining from scratch.
In the post-hoc setting~(\secref{end_to_end}), we only fine-tune the \transformation embeddings and train LoRA adapters on the final $k=8$ transformer blocks, leaving all other parameters frozen. Across five models and five languages, our method removes up to 10\% of the vocabulary tokens while largely maintaining performance over a suite of downstream tasks. 
In pretraining proof-of-concept experiments~(\secref{pretraining}), we show that compositional vocabularies are even more effective when trained from scratch: they remove 41\% of BPE vocabulary entries and obtain comparable performance, creating substantial room for new tokens.

In summary, we introduce compositional structure into language model vocabularies, enabling more efficient use of a fixed vocabulary budget through shared building blocks, which reduce redundancy in token allocation while also expanding lexical coverage. Our experiments demonstrate that LLMs can naturally operate over these representations, and establish compositional vocabularies as a competitive alternative to standard surface-form tokenization for future language models.

\section{Background: Token Allocation in Language Model Vocabularies}
\label{sec:motivation}

Tokenization bridges natural language and model representations: it decomposes text into sequences of tokens from a fixed vocabulary, where each token is an atomic string unit for which the model learns specialized, single-vector embeddings. These vocabularies are almost universally built using byte-pair encoding (BPE;~\citealp{bpe2016sennrich}), which iteratively merges the most frequent token pairs—from characters to subwords to words---in an attempt to optimally compress the text using a predetermined vocabulary size.

As LLM vocabularies grow larger~(e.g., \citetalias{gemma}; \citealp{aya23}), there is growing recognition that vocabulary resources can be better allocated. Recent studies point to stark imbalances in token allocations across languages, negatively impacting both model cost~\citep{petrov2023language, ahia2023languages} and performance~\citep{tokenizer-choice-2024, Limisiewicz2023TokenizationIM, Toraman2022ImpactOT}. These findings motivate the development of techniques for post-hoc vocabulary expansion to reduce costs for a specific language or domain~\citep{han2025adapters,nakash2025adaptivocab,liu-etal-2024-ofa,minixhofer2024zero}. 

Another line of research advocates for scaling up the vocabulary together with model size, to unlock performance gains in the model's main language~\citep{tao2024scaling, huang2025vocabscaling, superbpe2025}. Still, expansion is ultimately bounded by memory and compute constraints~\citep{dagan2024getting, wijmans2025cut}, underscoring the importance of carefully reconsidering how the token vocabulary is allocated.

\section{Word Structure and Redundancy in Vocabulary Design}
\label{sec:vocabulary_redundancy}
One underexplored source of inefficiency in current vocabulary design is the treatment of morphologically related word forms as independent tokens. In high-resource languages like English, this often results in large clusters of surface variants—\emph{walk, walks, walking, walked}—each assigned a separate token, despite their shared meaning and structure.

\begin{figure}[t]
   \centering
   \includegraphics[width=0.98\columnwidth]{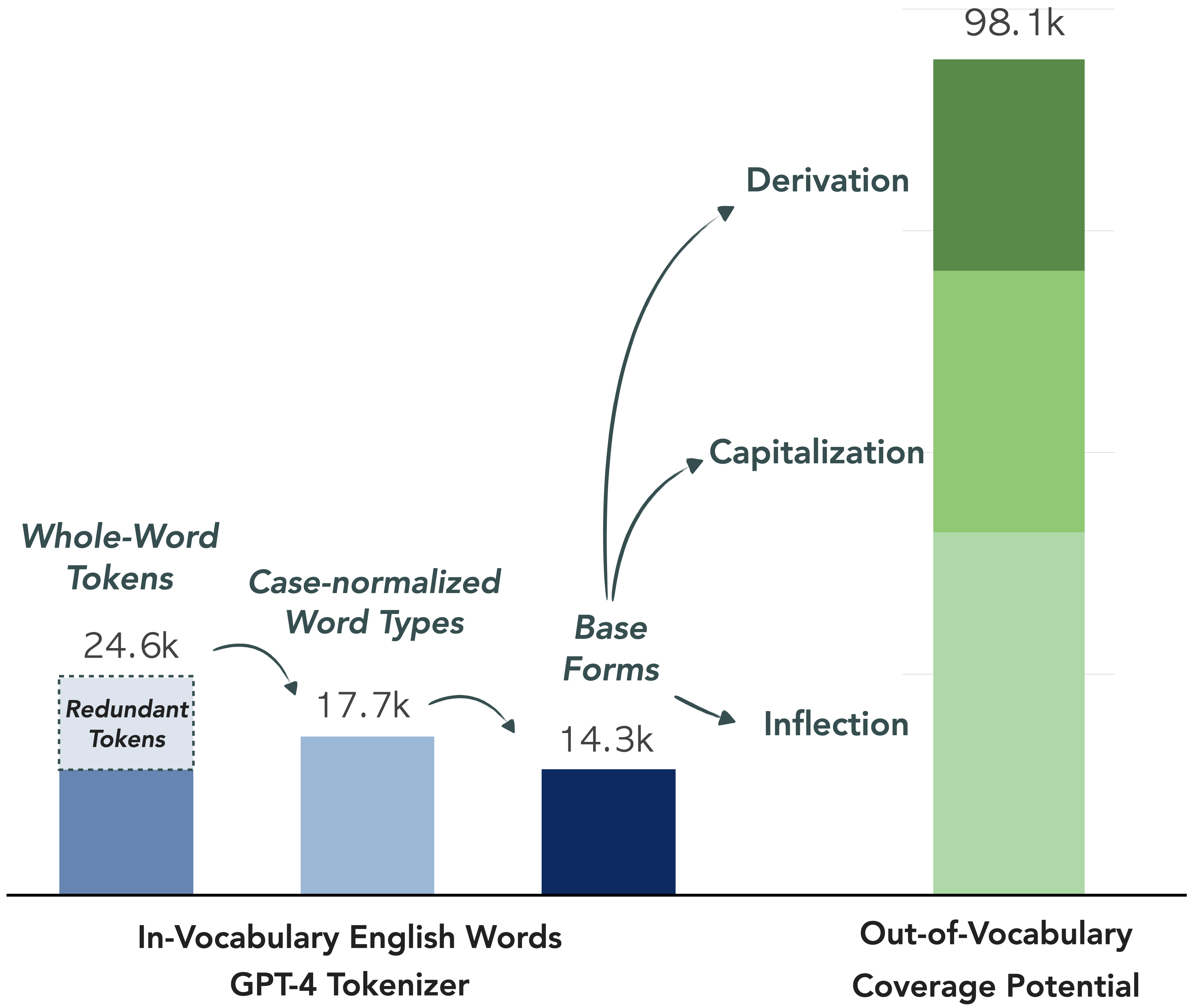}
   \caption{
\textbf{Structure in LLM vocabularies and potential for compositional design.} 
\textbf{Left:} Many in-vocabulary English word tokens in the GPT-4 tokenizer are surface variants of other tokens—differing only by case, inflection, or derivation—reducing from 24k tokens to just 14k base-form words. 
\textbf{Right:} The existing set of base forms and transformations can be used to compose over 98k currently out-of-vocabulary words, highlighting the inefficiencies of current vocabularies and the potential of a compositional design.}
\label{fig:motivation}
   
\end{figure}

To quantify this redundancy, we examine the English whole-word tokens in the GPT-4 tokenizer~\citep{openai2024tiktoken}---the base tokenizer for many recent LLMs~\citep{llama3,qwen2.5,olmo2}. We use UniMorph's English lexicon~\citep{batsuren-etal-2022-unimorph} to identify tokens that are English words,\footnote{We only consider tokens that start with a leading space as whole word tokens; tokens without it can sometimes occur mid-word (like ``ask'' in ``task'', compared to ``\_ask'').} finding 24.6k such tokens~(\figref{motivation}, left side).\footnote{Out of 100k tokens, there are 41.3k tokens with a leading space in the vocabulary that are composed of English letters. Roughly 60\% are identified as valid English words. The rest are either code-related terms, sub-words, or proper nouns. The other 60k tokens are either sub-words or non-English tokens.} Ignoring case (e.g., equating ``walk'' with ``Walk'') reduces this to 17.7k unique types. Further accounting for inflectional and derivational relations reduces this to just 14.3k \bases, a total reduction of 42\%.

Rather than assigning each word form a distinct, independently learned token, what if we could model these processes as \textsl{\transformations} applied to a compact set of base words? Our analysis shows that, beyond reconstructing every in-vocabulary word, these tokens can further represent 98k out-of-vocabulary words~(Fig.~\ref{fig:motivation}, right), which are currently represented using multiple tokens.

Altogether, this motivates a structured vocabulary design that composes word forms from shared blocks, yielding vocabularies that are simultaneously more compact and more expressive while scaling effectively across domains and languages.

\section{Composing Words from Base Forms and Transformations}
\label{sec:method}

We propose a compositional representation approach in which each surface form is constructed from a base word and a set of transformation vectors. Formally, let $\mathcal{V}_{\text{orig}}$ denote the  model’s original token vocabulary. We define a subset $\mathcal{V}_b \subset \mathcal{V}_{\text{orig}}$ as the base-word vocabulary, consisting of canonical lexical forms (e.g., \textit{walk}) and any auxiliary tokens (e.g., punctuation, sub-words, code segments, words in non-target languages). We also introduce a transformation vocabulary $\mathcal{V}_t$, which consists of a small number of vectors corresponding to morphological operations such as inflection or derivation, or other word-level processes like capitalization.

In our scheme, a word $w$ is represented by a base $b_w \in \mathcal{V}_b$ and a set of transformations $T(w) \subset \mathcal{V}_t$:
\begin{equation}\label{eq:1}
    \mathbf{e}_w = \mathbf{e}_{b_w} + \sum_{t_i \in T(w)} \mathbf{e}_{t_i}
\end{equation}
where $\mathbf{e}_{b_w}$ and $\mathbf{e}_{t_i}$ are rows from embedding matrices $E_b$ and $E_t$, respectively. We define the \emph{compositional vocabulary} $\mathcal{V}$ as all words that can be constructed from $(b_w, T(w))$ combinations. For base words and auxiliary tokens, $T(w) = \emptyset$.

This decomposition applies both at input and output: At input, we replace direct lookup with~\equref{1}. At output, we replace the model's large unembedding matrix $U$ with two separate matrices for \bases and \transformations: $U_b$ and $U_t$.
Given an output state $\mathbf{h}$, we score each candidate next-token $w$ by separately projecting $\mathbf{h}$ onto $U_b$ and $U_t$, and summing the relevant dot-products:
\begin{equation}\label{eq:2}
    \text{logit}(w) = \mathbf{h} \cdot \mathbf{u}_{b_w} + \sum_{t_i \in T(w)} \mathbf{h} \cdot \mathbf{u}_{t_i}
\end{equation}
where $\mathbf{u}_{b_w}$ and $\mathbf{u}_{t_i}$ are the corresponding columns of $U_b$ and $U_t$ for $w$'s components. To obtain the final next-token probabilities in the post-hoc setting, we apply a single softmax over the logits of all entries in $\mathcal{V}$~(as computed by \equref{2}).
Importantly, our method is agnostic to whether $w$ is originally in-vocabulary~(IV) or out-of-vocabulary~(OOV), as long as its base form is IV.

\paragraph{Vocabulary decomposition map.}
To apply this framework, we construct a mapping $w~\mapsto~(b_w,\;T(w))$ from surface forms to their base forms and matching transformations. We use  UniMorph~\citep{batsuren-etal-2022-unimorph}, a multilingual word form database, to identify base forms and their inflected and derived forms. Transformation labels are drawn from UniMorph’s standardized tags (e.g., \texttt{V;PST}) with added rules for capitalization. 
Then, to build a decomposition map for a given tokenizer's vocabulary $\mathcal{V}_{\text{orig}}$, we iterate over its tokens, identify base forms, and map all related surface forms---whether in-vocabulary or not---to their base and transformation sets.

Notably, the decomposition map could also be built using sources other than morphological annotations, such as unsupervised morphological segmentation~\citep{creutz-lagus-2002-unsupervised,creutz2007unsupervised,smit-etal-2014-morfessor, abdelali-etal-2016-farasa} or LLM-based morphological analyses~\citep{pranjic-etal-2024-llmsegm}. In this work, UniMorph serves as a clean experimental scaffold for testing whether models interpret and use these compositions correctly. Importantly, it is not a requirement of our framework itself.

\paragraph{Computing the transformation vectors.}
To initialize the transformation vectors themselves (i.e., the entries in $E_t$ and $U_t$) in already-trained models, we revisit the idea of vector arithmetic in embedding space~\citep{mikolov-etal-2013-linguistic-regularities}. Let $O$ be an embedding matrix of $\mathcal{V}_{\text{orig}}$, and let  $b(w):\mathcal{V}_{\text{orig}}\mapsto\mathcal{V}_{\text{b}}$ be a function that maps a word to its base form. For each transformation $t$, we extract the set $R(t) = \{\left(w, b(w)\right) \mid t\in T(w)\}$ of word pairs in $\mathcal{V}_{\text{orig}}$ that exemplify $t$ (e.g., \textit{walk} and \textit{walked} for $t=\text{past tense}$).\footnote{To obtain a ``clean'' signal for \transformations, we only use $w$ that demonstrate a \emph{single} transformation $(|T(w)|=1)$.} We then compute the average offset of their respective embeddings:
\begin{equation}\label{eq:3}
    \mathbf{o}_t = \frac{1}{|R(t)|}\sum_{w \in R(t) } (\mathbf{o}_w - \mathbf{o}_{b(w)})
\end{equation}
We compute this separately for all $t\in\mathcal{V}_\text{t}$ in both the embedding and unembedding spaces, yielding \transformation vectors for input and output.
While prior work analyzed such linearity in the embeddings of LLMs~\citep{park2024linearrepresentationhypothesisgeometry, park2025hierarchical}, to the best of our knowledge, our work is the first to leverage this for end-to-end language modeling.

\section{Do LLMs Understand Compositional Word Representations?}
\label{sec:does_it_work}

We now turn to our first core question: can LLMs that were pretrained with standard vocabularies interpret our compositional embeddings---sums of \base and \transformation vectors---as intended? 

Recent work has shown LLMs build up and resolve the meanings of input tokens across their early layers, a process referred to as \emph{detokenization}~\cite{kaplan2025from,feucht-etal-2024-token,gurnee2023finding}. This was particularly observed for multi-token words or in-vocabulary words split into multiple tokens (e.g., due to typos). Building on this, we feed models with compositional inputs and inspect whether the embedding and early layer representations have successfully resolved into the intended surface form meanings. 
To interpret these internal representations, we follow \citeauthor{kaplan2025from} and use \patchscopes~\citep{ghandeharioun2024patchscopes}, a prompting method to probe the contents of a hidden state using natural language.

\paragraph{Languages and models.}
We experiment with five morphologically-diverse languages: English, Arabic, German, Russian and Spanish. For English, we use three LLMs: \texttt{Llama-3-8B}~\citep{llama3}, \texttt{Qwen2.5-7B}~\citep{qwen2.5}, and \texttt{OLMo-2-7B}~\citep{olmo2}. As coverage of whole-word tokens in these models' vocabularies for other languages is narrow,\footnote{This restricts both the base-word lexicon, and the number of existing \textsl{base-inflected} pairs for extracting transformations.} we use models with dedicated tokenizers for them: \texttt{ALLaM-7B} for Arabic~\citep{bari2025allam} and \texttt{EuroLLM-9B} for the three other languages~\citep{martins2025eurollm}.\footnote{All models have vocabularies of 100k or more tokens, except \texttt{ALLaM} with 64k (but roughly 32k are for Arabic).} In experiments for a specific model and language pair, we construct the vocabulary decomposition and transformation vectors (\secref{method}) only for that language, ignoring words in other languages.

\paragraph{Examining word representations.}
For each model and language pair, we iterate over all words $w$ that could be composed from the \bases and \transformations extracted from its vocabulary~(\secref{method}).
Next, given a surface form $w$, we replace the token embedding for $w$ with its compositional representation $e_w$~(\equref{1}), and feed it to \patchscopes to generate its textual description.\footnote{Following \citet{kaplan2025from}, we use the \patchscopes prompt ``\texttt{[X], [X], [X], [X],}'', where we replace the placeholder token (\texttt{[X]}) with a hidden state $\mathbf{h}$ and let \patchscopes generate text. We expect \patchscopes to generate the intended word form if $\mathbf{h}$ indeed captures it. For languages other than English, we add the prefix ``\texttt{In \{language\_name\}:}''.} We then evaluate whether the \patchscopes interpretation of the compositional embedding $e_w$ matches the target word~$w$~(\emph{embed}).
We also examine whether the model successfully \emph{detokenizes} compositional embeddings in its early layers: we feed $e_w$ to the model without any context, extract the resulting hidden states at the first $k=10$ layers, and report whether the \patchscopes interpretation matches the target word $w$ in at least one layer~(\emph{detok}).

\paragraph{English results.}
We begin by examining English words that exist as single tokens in \texttt{Llama-3-8B}'s original vocabulary $\mathcal{V}_{\text{orig}}$~(\tabref{patchscopes_Llama_3.1_8B}, \emph{in-vocab}). We observe that most inflectional transformations---such as verb tense (past, present participle) and number (plural)---as well as capitalization, are often correctly resolved by the model already at the embedding layer~(\emph{embed}), and almost always at early internal layers~(\emph{detok}). For example, $\mathbf{e}_\text{walk}+\mathbf{e}_\textsl{past}$ is interpreted by \patchscopes as ``\textit{walked}''. 
In contrast, derivations (e.g., \textit{walk}$\to$\textit{walkable}), which rarely occur as single tokens in the vocabulary,  are seldom recognized by the model and often resolve as the base word instead. This suggests that models learn weaker linear structure for rare relations, or that \transformation vectors built using small sample sizes show weaker generalization.

We next examine out-of-vocabulary words, i.e., English words that can be composed using the \bases and \transformations but are \emph{not} found as a single token in the original vocabulary~(\tabref{patchscopes_Llama_3.1_8B}, \emph{out-of-vocab}). Using our decomposition map, we construct single-vector representations for these words and feed them to the model. Surprisingly, many of these are resolved as the intended word form already at the embedding layer, with \patchscopes generating the full, multi-token word, especially for inflections and capitalization. Similarly to in-vocabulary results, we observe higher successful resolution rates for early-layer detokenization,~while representing out-of-vocabulary derivations compositionally generally fails. We observe similar results for English in other models.\footnote{See~\appref{appendix_patchscopes_english_results} for further English results, and \appref{appendix_patchscopes_analysis} for analysis of \transformation errors and geometry.}

\begin{table}[!t]
\centering
\resizebox{0.98\columnwidth}{!}{%
\begin{tabular}{l@{\hspace{0.5em}}ccc@{\hspace{0.5em}}ccc}
\toprule
\textbf{Transformation} & \multicolumn{3}{c}{In-vocab.} & \multicolumn{3}{c}{Out-of-vocab.} \\
\cmidrule(lr){2-4} \cmidrule(lr){5-7}
 & \emph{embed} & \emph{detok} & $N$ & \emph{embed} & \emph{detok} & $N$ \\
\midrule
\multicolumn{7}{l}{\textbf{Inflection}} \\
\hspace{1em}Plural (N) & 92\% & 96\% & 0.8k & 30\% & 56\% & 3.4k \\
\hspace{1em}\shortstack[l]{Plural (N) \\\hspace{0.5em} \& Present Singular (V)} & 87\% & 91\% & 1.6k & 43\% & 75\% & 2.1k \\
\hspace{1em}Present Singular (V) & 90\% & 91\% & 0.1k & 64\% & 82\% & 0.3k \\
\hspace{1em}Past (V) & 71\% & 81\% & 0.6k & 9\% & 29\% & 2.9k \\
\hspace{1em}Past Participle (V) & 64\% & 93\% & 14 & 14\% & 38\% & 21 \\
\hspace{1em}Gerund (V) & 83\% & 93\% & 0.2k & 17\% & 34\% & 3.2k \\
\hspace{1em}Superlative (ADJ) & 71\% & 94\% & 31 & 5\% & 29\% & 0.4k \\
\hspace{1em}Comparative (ADJ) & 40\% & 83\% & 30 & 3\% & 36\% & 0.4k \\
\midrule
\textbf{Capitalization} & 80\% & 89\% & 6.0k & 72\% & 85\% & 8.4k \\
\midrule
\multicolumn{7}{l}{\textbf{Derivation}} \\
\hspace{1em}-y & 24\% & 47\% & 17 & 2\% & 12\% & 1.5k \\
\hspace{1em}-er & 8\% & 17\% & 12 & 0\% & 6\% & 2.6k \\
\hspace{1em}-al & 25\% & 25\% & 8 & 0\% & 9\% & 0.7k \\
\hspace{1em}un- & 0\% & 33\% & 3 & 0\% & 2\% & 3.3k \\
\hspace{1em}re- & 67\% & 67\% & 3 & 0\% & 10\% & 1.8k \\
\hspace{1em}-ic & 100\% & 100\% & 2 & 4\% & 21\% & 0.4k \\
\hspace{1em}All derivatives & 31\% & 45\% & 51 & 0\% & 3\% & 31.4k \\
\bottomrule
\end{tabular}
}
\caption{Accuracy of \patchscopes interpretations for compositional input representations (i.e., \base + \transformation embeddings) of in-vocabulary and out-of-vocabulary English words in \texttt{Llama-3.1-8B}. We report successful resolution both at the embedding layer~(\emph{embed}), and after detokenization in early layers~(\emph{detok}). $N$ indicates the number of surface forms evaluated per category. Compositional embeddings of capitalization and inflectional forms are very often resolved correctly---even for many out-of-vocabulary words, which never occur as single input vectors during pretraining. Derivatives remain challenging---likely because they rarely occur as in-vocabulary words.}
\label{tab:patchscopes_Llama_3.1_8B}
\end{table}

\begin{table*}[!t]
\centering
\resizebox{0.98\textwidth}{!}{%
\begin{tabular}{l@{\hspace{0.5em}}l@{\hspace{0.5em}}cc@{\hspace{0.5em}}cc@{\hspace{0.5em}}cc@{\hspace{0.5em}}cc@{\hspace{0.5em}}cc}
\toprule
& \multirow{2}{*}{{Language}} & \multicolumn{2}{c}{{Capitalization}} & \multicolumn{2}{c}{{Noun Inflection}} & \multicolumn{2}{c}{{Adjective Inflection}} & \multicolumn{2}{c}{{Verb Inflection}} & \multicolumn{2}{c}{{Derivation}} \\
\cmidrule(lr){3-4} \cmidrule(lr){5-6} \cmidrule(lr){7-8} \cmidrule(lr){9-10} \cmidrule(lr){11-12}
 & & \emph{In-Vocab.} & \emph{Out-Vocab.} & \emph{In-Vocab.} & \emph{Out-Vocab.} & \emph{In-Vocab.} & \emph{Out-Vocab.} & \emph{In-Vocab.} & \emph{Out-Vocab.} & \emph{In-Vocab.} & \emph{Out-Vocab.} \\
\midrule
\emph{ALLaM} & \hspace{0.5em}Arabic & --- & --- & 77\% (1.8k) & 14\% (3.6k) & 69\% (0.5k) & 23\% (1.0k) & 41\% (1.0k) & 14\% (2.7k) & --- & --- \\ \addlinespace[0.4em]
\emph{EuroLLM} & \hspace{0.5em}German & 95\% (0.2k) & 74\% (0.4k) & --- & --- & 21\% (0.3k) & \phantom{0}7\% (1.3k) & 82\% (0.3k) & 36\% (1.2k) & --- & --- \\
 & \hspace{0.5em}Russian & 97\% \phantom{.k}(66) & 88\% (0.7k) & 63\% (0.6k) & 21\% (4.2k) & 100\% \phantom{.k}(50) & 89\% \phantom{.k}(94) & 83\% \phantom{0.k}(6) & 30\% \phantom{.k}(10) & --- & --- \\
 & \hspace{0.5em}Spanish & 97\% (1.0k) & 90\% (2.8k) & 76\% (0.7k) & 46\% (1.9k) & 79\% (0.5k) & 60\% (1.1k) & 67\% (0.8k) & 35\% (6.9k) & 37\% (65) & 14\% \phantom{1}(0.4k) \\ \addlinespace[0.5em]
\emph{Llama-3} & \hspace{0.4em}English & 80\% (6.0k) & 72\% (8.4k) & 89\% (2.4k) & 35\% (5.6k) & 56\% (61) & \phantom{0}4\% (0.9k) & 76\% (0.9k) & 16\% (6.4k) & 20\% (41) & 0\% (12.8k) \\
\bottomrule
\end{tabular}
}
\caption{Accuracy of \patchscopes interpretations for compositional input embeddings across languages. Numbers in parentheses indicate sample sizes. "---" indicates cases where no suitable \textsl{base-inflection} pairs were found in the vocabulary or where there are no UniMorph entries for that category. For detokenization results, see~\appref{appendix_patchscopes_embedding}.}
\label{tab:multilingual_patchscopes}
\end{table*}

\paragraph{Multilingual results.}
We repeat the same experiment on each of the other languages. Since each language has different types and number of inflectional and derivational processes,\footnote{We treat each UniMorph tag as its own \transformation.} we aggregate results over five categories: adjective inflection, verb inflection, noun inflection, derivation and capitalization. Our results~(\tabref{multilingual_patchscopes}) show that LLMs can correctly interpret compositional word representations across diverse languages and morphological structures. Surprisingly, some \transformation vector types (e.g., adjective or verb inflections) work better for out-of-vocabulary representation than in English, hinting that models learn stronger linear encodings of morphological structure when the token vocabulary is more limited---a phenomenon we further analyze in~\secref{analysis}. 
Overall, our results show that LLMs can naturally interpret compositional word embeddings across languages.

\paragraph{Analysis of composition failures.}
Across languages and models, we observe a consistent gap between inflectional transformations (often resolved) and derivational transformations (rarely resolved). To characterize these failures, we analyze whether the number of in-vocabulary exemplar pairs used to estimate each \transformation vector (\equref{3}) helps explain composition failures. We find that the number of exemplars mainly matters for \emph{generalization}: \transformation vectors estimated from many pairs are much more likely to resolve for multi-token surface forms, while in-vocabulary success is overall insensitive to exemplar count once a usable signal is available (see \appref{exemplar_count_correlations}).

\section{Compositional Language Modeling}
\label{sec:end_to_end}
We have shown that \transformation vectors capture meaningful operations in the input space of LLMs, and that these can be successfully composed with base word embeddings. We next investigate whether models can use compositional vocabularies effectively in end-to-end language modeling.

\subsection{Implementation and Experimental Setup}
\label{subsec:implementation}
Given a model's vocabulary decomposition map~(\secref{method}), we apply our compositional vocabulary framework and restructure the input and output embedding matrices. We replace the model's input embedding of any surface form $w$ with compositions of the corresponding \base and \transformation embeddings~(\equref{1}). For next-token prediction, we compute logits through summation of \base and \transformation logits~(\equref{2}). Importantly, any word not in the decomposition map maintains its original embedding and unembedding throughout training and inference, without modifications.

\paragraph{Fine-tuning the transformation vectors.} After initialization~(\equref{3}), we train the \transformation vectors jointly within the model: we treat the \transformation embedding and unembedding matrices $E_t$ and $U_t$ as trainable weights (introducing fewer than 0.001\% additional parameters), and freeze all other model parameters, including the embeddings and unembeddings of \bases. We use knowledge distillation loss~\citep{hinton2015distilling} to fine-tune the \transformation vectors using two-stage distillation: We first freeze the output unembeddings and only train the \textsl{input} \transformations, using the predictions of the original, unmodified model as targets. Next, we freeze the input embeddings and only train the \textsl{output} \transformations, this time using the (frozen) model resulting from the first stage as the distillation target---ignoring all words $w\notin \mathcal{V}_{\text{orig}}$ in the loss. In both stages, we train on a fixed, small sample of the \texttt{FineWeb-Edu} corpus~\citep{penedo2024fineweb}.\footnote{We use a sequence length of 256 and train on $\sim$5M tokens.} See~\appref{appendix_implementation}.

\paragraph{Lightweight LoRA adaptation.}
To allow lightweight adaptation to the reshaped output vocabulary, we add LoRA adapters to the final $k=8$ model layers, keeping all other internal layers frozen. We use LoRA $r=\alpha=256$.

\paragraph{Filtering the decomposition map.}
Our results in~\secref{does_it_work} indicate some out-of-vocabulary surface forms fail to be interpreted by the model as their intended word when given as compositions. We therefore filter out surface words with failed detokenization from the decomposition map, and fall back to using their original tokenization and embeddings in both input and output. We also exclude all derivational transformations due to their weak resolution rates.  See analysis in \appref{appendix_filtering}.

\paragraph{Downstream tasks.} We evaluate our compositional vocabulary models on a diverse suite of standard benchmarks. As a baseline, we compare performance to the original, unmodified models. For English, the benchmarks cover knowledge, reading comprehension, and commonsense: \textsl{MMLU}~\citep{hendryckstest2021}, \textsl{ARC}~\citep{clark-etal-2018-think}, \textsl{HellaSwag}~\citep{zellers-etal-2019-hellaswag}, \textsl{Winogrande}~\citep{sakaguchi2021winogrande}, \textsl{TriviaQA}~\citep{joshi-etal-2017-triviaqa}, \textsl{SQuAD}~\citep{rajpurkar-etal-2016-squad}, \textsl{BoolQ}~\citep{clark-etal-2019-boolq}, \textsl{PIQA}~\citep{Bisk2020} and \textsl{COPA}~\citep{kavumba-etal-2019-choosing}. For other languages, we use \textsl{XNLI}~\citep{conneau-etal-2018-xnli}, \textsl{XQuAD}~\citep{artetxe-etal-2020-xquad} and \textsl{Global MMLU}~\citep{singh-etal-2025-globalmmlu}. 
See \appref{appendix_eval}.

\subsection{Results}
\label{sec:lm_results}
We report our results for English on \texttt{Llama-3-8B} in~\tabref{downstream_Llama_3.1_8B},\footnote{See~\appref{appendix_posthoc_results} for results on other English models.} and results for other languages in~\tabref{multilingual_results}. 
Our compositional language modeling approach results in minimal degradation compared to the baseline models across languages, indicating that LLMs can leverage compositional vocabularies with only lightweight adaptation. 

We further inspect reductions in vocabulary size after applying our framework. For English, our approach removes roughly 10k surface-form tokens from \texttt{Llama3} and \texttt{OLMo2} each, and 7.8k from \texttt{Qwen2.5}.\footnote{In other languages, absolute reductions are smaller (0.6k--3k) but correspond to 38--45\% of whole-word tokens in the target languages, as these tokenizers devote far fewer whole-word entries to non-English languages to begin with.} This frees a meaningful number of vocabulary slots for reallocation: recent work has shown that adding even several hundred dedicated tokens to the vocabulary can greatly improve tokenization efficiency and downstream behavior for a language or expert domain~\citep{ahia2023languages,liu-etal-2024-gold,nakash2025adaptivocab}. 
Notably, our method has a marginal effect on decoding speed---only a 0.8\% reduction compared to standard prediction~(see \appref{appendix_decoding_speed}). 

\paragraph{Reallocating freed vocabulary slots.} To make the practical gains concrete, we simulate token reallocation based on our results for the \texttt{Llama-3.1-8B} tokenizer. After evicting the 10k English surface words that can be represented compositionally, we add 2.5k new language-specific BPE tokens for each of: Arabic, Russian, German, and Spanish. We measure compression using bytes-per-token (BPT) on held-out \texttt{FineWeb-2}~\citep{fineweb2} text. After reallocation, BPT improves from 4.40 to 4.81 on average across languages~(\tabref{reallocation_bytes_per_token}).

In the next section, we show that vocabularies can also be built compositionally from the outset, with even greater vocabulary-allocation efficiency, by pretraining a model with a compositional vocabulary from scratch (\secref{pretraining}).

\begin{table}[!t]
\centering
\resizebox{0.98\columnwidth}{!}{%
\begin{tabular}{llcccc}
    \toprule
        \textbf{Category} & \textbf{Task} & \textbf{Baseline} & \textbf{End-to-end} & \textbf{$\Delta$} \\\midrule
       Knowledge & MMLU \textsubscript{(Acc.)} & 65.2 & 64.9 & -0.3 \\
        & ARC \textsubscript{(Acc.)} & 53.6 & 52.5 & -1.1 \\
        \midrule
       \multirow{2}{*}{\makecell[lt]{Reading\\\hspace{0.5em}Comprehension}} & BoolQ \textsubscript{(Acc.)} & 83.2 & 83.3 & +0.1 \\
        & TriviaQA \textsubscript{(EM)} & 66.5 & 63.3 & -3.3 \\
        & SQuAD \textsubscript{(EM)} & 22.1 & 20.0 & -2.1 \\
        \midrule
       Commonsense & Hellaswag \textsubscript{(Acc.)} & 60.6 & 59.5 & -1.1 \\
        & Winogrande \textsubscript{(Acc.)} & 78.1 & 78.6 & +0.5 \\
        & PIQA \textsubscript{(Acc.)} & 80.3 & 79.1 & -1.2 \\
        & COPA \textsubscript{(Acc.)} & 93.0 & 92.0 & -1.0 \\
        \midrule
         Average & & \textbf{66.9} & \textbf{65.9} & {-1.0} \\
    \bottomrule
    \end{tabular}

}
\caption{Downstream performance of English compositional-vocabulary models~(\textsl{End-to-end}) and their original, unmodified version~(\textsl{Baseline}) for \texttt{Llama-3.1-8B}.
Our framework remains competitive with the baseline despite extensive changes to the model's input and output representation mechanisms---highlighting the intrinsic ability of LLMs to process and predict words compositionally.}

\label{tab:downstream_Llama_3.1_8B}
\end{table}

\begin{table}[!t]
\centering
\resizebox{0.98\columnwidth}{!}{%

\begin{tabular}{l@{\hspace{0.7em}}l@{\hspace{1em}}c@{\hspace{0.3em}}c@{\hspace{1.5em}}c@{\hspace{0.3em}}c@{\hspace{1.5em}}c@{\hspace{0.3em}}c}
    \toprule
          & & XNLI & $\Delta$ & XQuAD & $\Delta$ & GMMLU & $\Delta$ \\

        \midrule
\emph{ALLaM} & Arabic & 44.1 & -0.3 & 42.7 & -3.2 & 59.9 & +0.2 \\ \addlinespace[0.3em]
\emph{EuroLLM} & German & 46.5 & +0.6 & 51.3 & -1.6 & 54.6 & -0.7 \\
 & Russian & 40.1 & -4.5 & 37.4 & -3.6 & 54.4 & -0.3 \\
 & Spanish & 43.3 & -0.6 & 48.3 & -4.1 & 55.2 & -0.9 \\ 
    \bottomrule
    \end{tabular}
}
\caption{Multilingual downstream performance of compositional-vocabulary models, along with absolute performance difference from the baseline model ($\Delta$).}
\label{tab:multilingual_results}
\end{table}

\section{Compositional Vocabulary Pretraining}
\label{sec:pretraining}
To demonstrate that compositional vocabularies can also serve as a \emph{design choice} when training new language models, we reshape English and Spanish BPE vocabularies into compositional ones, and pretrain small baseline and compositional models from scratch. For English, we reshape the 50k-token \texttt{GPT-2} tokenizer~\cite{gpt2}, while restricting the compositional model to predict exactly the same surface-form vocabulary as the BPE baseline (i.e., we do not extend to out-of-vocabulary words). For Spanish, we train a 32k-token BPE vocabulary\footnote{We train the Spanish tokenizer on 10B bytes from the Spanish subset of FineWeb-2~\cite{fineweb2}} and then reshape it, this time allowing the compositional model to generate out-of-vocabulary surface forms via compositions. For each language and vocabulary, we pretrain a \texttt{nanoGPT-124M} model~\cite{modded_nanogpt_2024} on 1B tokens,\footnote{We use FineWeb (English) and FineWeb-2 (Spanish).} comparing a baseline model against an otherwise-identical compositional model. 

In contrast to our post-hoc setup, the compositional model predicts tokens in a factorized space, where a surface word $w$ is predicted by first sampling from a \base distribution, and then predicting \transformations conditioned on a chosen \textsl{base}:
\begin{equation}
    p(w \mid \mathbf{h}) = p(b_w \mid \mathbf{h})\, p(T(w) \mid b_w, \mathbf{h})
\end{equation}
We further include a space-prefix \transformation~(e.g., ``\_\textit{walking}'' vs.\ ``\textit{walking}'').\footnote{Modern BPE vocabularies include prefix whitespace characters when merging tokens, creating many near-duplicates.}
We measure performance using bits-per-byte (BPB) on a held-out set, as it is well-defined across different vocabularies and tokenizers.\footnote{BPB normalizes negative log-likelihood by the number of UTF-8 bytes in the evaluation text.} To measure tokenization efficiency in Spanish, we use average bytes-per-token (higher is better). See \appref{appendix_pretraining} for the exact training and modeling details.

\begin{table}[t]
\centering
\resizebox{0.72\columnwidth}{!}{
\begin{tabular}{lcccc}
\toprule
Language & Baseline & Reallocated & $\Delta$ (\%) \\
\midrule
Arabic & 4.62 & 5.46 & +18.0 \\
Russian & 5.59 & 5.85 & +4.8 \\
German & 3.59 & 3.86 & +7.5 \\
Spanish & 3.80 & 4.07 & +7.0 \\
\midrule
Average & 4.40 & 4.81 & +9.3 \\
\bottomrule
\end{tabular}
}
\caption{Tokenization efficiency, measured in bytes-per-token (\emph{higher is better}), before and after reallocating token slots with our approach. Starting from the \texttt{Llama-3.1-8B} tokenizer, we replace 10k English surface forms that are represented compositionally with 2.5k new, non-overlapping BPE tokens for each language, keeping the total vocabulary size fixed.}
\label{tab:reallocation_bytes_per_token}
\end{table}

\begin{table}[t]
\centering
\setlength{\tabcolsep}{4pt}
\resizebox{0.85\columnwidth}{!}{
\begin{tabular}{@{}lccccc@{}}
\toprule
& & \multicolumn{2}{c}{BPB $\downarrow$} & \multicolumn{2}{c}{Bytes/tok. $\uparrow$} \\
\cmidrule(lr){3-4} \cmidrule(lr){5-6}
Language & Vocab. red. & Base & Comp. & Base & Comp. \\
\midrule
English & 41.6\% & 1.08 & 1.09 & -- & -- \\
Spanish & 41.8\% & 1.00 & 1.11 & 4.77 & 4.92 \\
\bottomrule
\end{tabular}
}
\caption{Pretraining results for baseline (\emph{Base}) and compositional (\emph{Comp.}) models based on the same BPE vocabulary. Lower bits-per-byte (BPB) is better; higher bytes-per-token indicate more efficient tokenization. For Spanish, we further extend the compositional model to previously out-of-vocabulary word compositions, resulting in better compression.}
\label{tab:pretraining_summary}
\end{table}

We report our results in \tabref{pretraining_summary}. In both languages, our approach frees roughly 42\% of vocabulary entries  compared to the original tokenizers. English shows comparable performance under this more compact parameterization, whereas Spanish shows a small BPB gap alongside more efficienct tokenization---while using an overall much smaller vocabulary.

Together, these results show that compositional vocabularies can be trained from scratch effectively, offering compact vocabularies and improved tokenization efficiency for future language models.

\section{Morphology in Embedding Space Scales Inversely with Vocabulary Size}
\label{sec:analysis}
\label{appendix_vocab_scaling}

\begin{figure*}[t]
   \centering
   \includegraphics[width=0.95\textwidth]{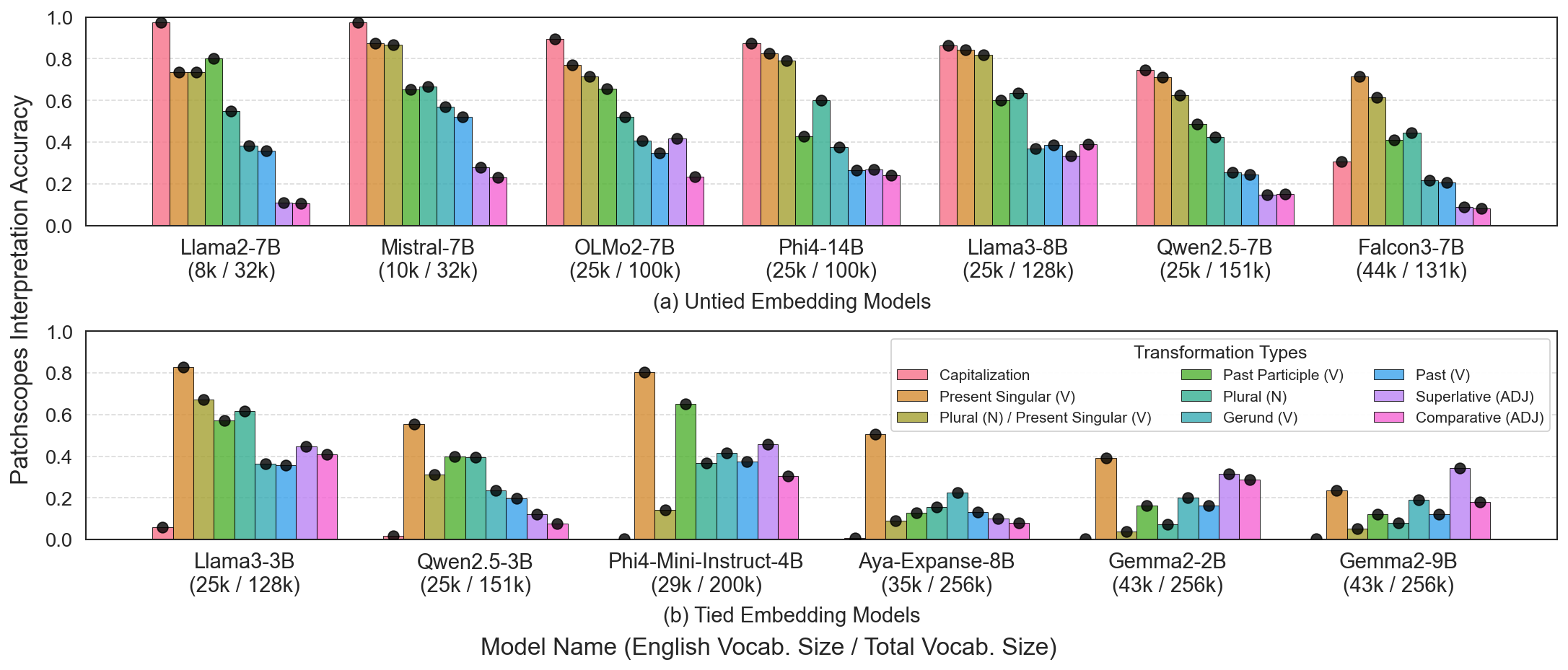}
   \caption{
\textbf{Linear representation of morphology in embeddings weakens as vocabulary size increases.} 
Accuracy of \patchscopes interpretations of compositional word representations across models, in order of increasing English vocabulary size (English tokens present in UniMorph), separated by embedding architecture. Scaling vocabulary size leads models to represent inflected forms as individual lexical units, rather than with consistent vector offsets.
}
\label{fig:vocab_scaling_analysis}
   
\end{figure*}

Having established that models implicitly learn compositional word representations, and with recent calls to scale vocabularies even further, a natural question emerges: how does vocabulary size affect the way models encode linguistic structure?

To study this question, we evaluate the extent of compositional word representations across models with varying vocabulary sizes. For each model, we decompose its vocabulary and measure the average \patchscopes interpretation accuracy for each \transformation vector we extract (as in \secref{does_it_work}). We also separate models by their embedding architecture~(\emph{untied} vs.~\emph{tied}). We track each model's English vocabulary size (the subset of tokens present in English UniMorph), and plot the results in order of increasing vocabulary size. The English vocabulary size of these models spans 8k--44k tokens  with total vocabulary sizes of 32k--256k tokens, representing varied scales of vocabulary design. See full model details in \appref{appendix_inverse_scaling}.

Our results~(\figref{vocab_scaling_analysis}) reveal a general inverse relationship: models with compact English vocabularies (8--10k words, e.g., \texttt{Llama2}, \texttt{Mistral}) tend to encode morphology through consistent vector offsets that generalize across words. In contrast, large-vocabulary models ($\sim$40k words, e.g., \texttt{Falcon3}, \texttt{Gemma2-9B}) tend to represent inflected forms of the same type as individual lexical units, rather than through a shared  linear translation of their base forms, with weight tying further amplifying this trend.
Overall, these results suggest that vocabulary scaling trades morphological compositionality in embedding space for lexical memorization.\footnote{Importantly, this does not imply that large-vocabulary models lack morphological knowledge, only that they rely less on linearly encoded morphology in their embedding space.}

\section{Related Work}

\paragraph{Incorporating morphology into representations}
A longstanding goal in NLP has been to integrate morphological knowledge into models. Early work on Transformer language models explored injecting linguistic features post-hoc~\citep{hofmann-etal-2021-superbizarre,gan2022morphte} or during pretraining~\citep{park-etal-2021-morphology,cui2022lert,matthews2018using,blevins2019better,hofmann-etal-2020-dagobert,seker-etal-2022-alephbert,peng-etal-2019-palm}, but such approaches are absent in modern LLMs. Recent work examined word segmentation effects on performance~\citep{marco-fraser-2024-subword,lerner-yvon-2025-unlike}, as well as morphology-aware tokenization to better reflect word structure~\citep{bauwens-delobelle-2024-bpe,asgari2025morphbpe}. Rather than injecting linguistic structure, we leverage compositional representations already present in LLMs.

\paragraph{Vector arithmetic of word representations}
Linear structure in word representations was first observed in Word2Vec~\citep{Mikolov2013DistributedRO,mikolov-etal-2013-linguistic-regularities,Levy2014LinguisticRI,Vylomova2015TakeAT}. Recent work found similar structures in LLMs across the unembedding layer~\citep{park2024linearrepresentationhypothesisgeometry,park2025hierarchical}, residual stream~\citep{merullo2023language,hendel-etal-2023-context,todd2024function}, and in behavior-steering directions~\citep{subramani-etal-2022-extracting,hernandez2024inspecting}. We further show that such structure is usable for end-to-end language modeling. 
Beyond morphology, \transformation vectors could capture semantic relations (e.g., country–nationality; \citealp{gladkova2016analogy}) or tie word embeddings across languages~\citep{schut2025multilingual}.

\paragraph{Post-hoc vocabulary modification}
Recent work has proposed methods to expand or modify token vocabulary by training new embeddings and fine-tuning internal model layers~\citep{kim2024efficient,takase2024large,han2025adapters,minixhofer2024zero,benartzy2025spellm,dobler-de-melo-2023-focus}. We avoid continual pretraining of model weights, and represent new forms by using the model's existing linguistic knowledge.

\section{Conclusion}

We have shown that word representations in LLMs are inherently compositional, and leveraged this property to introduce compositional vocabularies. Such vocabularies are more compact in size and more expressive in lexical coverage---freeing token slots that can be reallocated to words, languages, and domains that are currently tokenized inefficiently. Our results demonstrate that by integrating compositional vocabularies into future models, LLMs could cover more words, languages, and domains, without sacrificing performance.

\section*{Limitations}

Our framework employs external morphological resources to define transformation pairs. While this allows for clean experimental control, it limits immediate applicability to languages or domains lacking annotated morphological data. However, UniMorph serves as an experimental scaffold in this paper, not as a requirement of the framework itself: in principle, the method only needs a decomposition map, which could come from unsupervised segmentation, statistical morphology learning, or bootstrapped analyses. In future work, we will explore whether transformation vectors can be induced directly from data in an unsupervised fashion.

Post-hoc adaptation is bounded by what pretrained models already encode reliably enough for linear composition. This is most visible in derivational morphology, where composed input representations are often interpreted as the base form. Importantly, we do not observe such issues when pretraining models with compositional vocabularies from scratch.

Our vocabulary reshaping approach also assumes a relatively simple decomposition of each surface form into a base word and a set of transformation vectors. While effective for many cases, this simplification does not account for certain words which admit multiple plausible morphological analyses. Still, these problems are also encountered with standard tokenization approaches, with models learning to disambiguate such words into their intended meanings.

\section*{Acknowledgments}
We thank Guy Peskin and Amit Ben-Artzy for valuable conversations about this work. We are also grateful to the reviewers for their constructive and thoughtful feedback. This work was supported in part by the Israel Science Foundation (grant no. 2045/21) and by NSF-BSF grant 2020793.

\bibliography{custom,anthology}

\clearpage

\appendix
\section{Supplementary Results}
\label{appendix_results}

\subsection{English Patchscopes Results}
\label{appendix_patchscopes_english_results}
For the results of the \patchscopes experiments on other models, see \tabref{patchscopes_Qwen} and \tabref{patchscopes_OLMo}. 

\subsection{Multilingual Patchscopes Results}
\label{appendix_patchscopes_embedding}
For the results on multilingual \patchscopes interpretations of compositional input embeddings after detokenization, see \tabref{multilingual_patchscopes_detokenization}.

\subsection{Post-hoc Adaptation Results}
\label{appendix_posthoc_results}
For post-hoc adaptation results on other models, see \tabref{downstream_qwen} and \tabref{downstream_olmo}.

\section{Additional Analysis}
\label{appendix_analysis}

\subsection{Analyses of Patchscopes Results}
\label{appendix_patchscopes_analysis}

\paragraph{Patchscopes error breakdown.} We classify each \patchscopes generation of \texttt{Llama-3.1-8B} for English single-token \textsl{base}$+$\transformation targets into four outcomes: exact match of target, exact match of base form, exact match different inflection of the same base, and other~(\tabref{appendix_patchscopes_errors}). The dominant error is collapse to the base form---showing the model either interprets the compositional representation correctly, or as the corresponding base word. We also verify that when the target is already a base form, \patchscopes almost always returns that same base form rather than an inflected variant.

\paragraph{Geometry of offset vectors.} Our \patchscopes and end-to-end language modeling results indicate that language models often represent words compositionally. We further compare each pairwise offset (e.g., \textit{walked}-\textit{walk}) to its own \transformation category vector versus other \transformation vectors using cosine similarity for \texttt{Llama-3.1-8B}. The resulting separation is clear in both input and output spaces (\tabref{appendix_offset_geometry_summary}): offsets are consistently closer to their own \transformation type than to others, and top-1 \transformation accuracy is high, indicating separability between \transformations. Still, while models \emph{functionally} operate as if each \transformation vector as a single crisp direction in embedding space (as shown in the \patchscopes and end-to-end language modeling experiments), these results indicate this is an over-simplification.

\subsection{Exemplar Count Predicts Out-of-Vocabulary Generalization for Transformation Vectors}
\label{exemplar_count_correlations}

We test whether the number of in-vocabulary exemplar pairs used to estimate each \transformation vector~(\equref{3}) predicts whether the resulting composed embedding is interpreted as the intended surface form. For each \transformation $t$, we use the number of single-token in-vocabulary (IV) base/surface pairs as a proxy for exemplar set size, and compute Spearman correlations with additive success on IV targets and on out-of-vocabulary (OOV; multi-token) targets separately.

For \texttt{Llama-3.1-8B}, across individual transformations ($n=24$), IV additive success is nearly independent of exemplar count (Spearman's $\rho=0.04$). By contrast, IV exemplar count strongly predicts OOV performance ($\rho=0.77$, $p=1.5\times 10^{-5}$). This dissociation suggests that exemplar richness is not the main bottleneck for IV targets, which largely saturate once a direction is available, but is important for generalization to multi-token OOV surface forms.

Restricting the analysis to transformations within each coarse class yields the same qualitative pattern, though with limited power due to small $n$. Among inflectional transformations ($n=8$), IV success correlates moderately with the number of IV exemplars ($\rho=0.50$). Among derivational transformations ($n=14$), IV success instead trends negative ($\rho=-0.44$), consistent with the observation that derivations remain difficult even when many IV exemplars are available. In both classes, OOV success remains positively correlated with IV exemplar count (inflection: $\rho=0.38$; derivation: $\rho=0.38$), suggesting that OOV generalization depends on having enough IV pairs, even if this alone does not close the gap.

\subsection{Filtering the Vocabulary Decomposition for Failed Surface Form Compositions}
\label{appendix_filtering}

In \secref{does_it_work} we have seen that, even though the compositional embeddings work well for many in- and out-of-vocabulary words, there are also failure cases where we cannot be certain that the model interprets the compositional representation correctly. Intuitively, this means that using these representations in end-to-end language modeling might hurt model performance; indeed, when we remove the surface forms corresponding to these failures from the decomposition map (and after fine-tuning the transformation vectors as usual), we observe an average 1.6 points improvement across downstream benchmarks, compared to no filtering. We note that for input-only restructuring, we observe no effect, likely because the model has more error-correction opportunities across its layers. We therefore apply this filtering in experiments in \secref{end_to_end}.

\subsection{Decoding Speed}
\label{appendix_decoding_speed}
Our compositional language modeling approach introduces some additional complexity into next-token prediction: to compute token scores over the full, extended vocabulary, we map and sum up logit contributions from the \base and \transformation vocabularies~(\equref{2}). To validate that this does not introduce meaningful overhead, we let both the baseline and compositional \texttt{Llama-3-8B} models generate text in response to prompts from the CNN-DailyMail dataset~\citep{chen-etal-2016-thorough}, and measure the average number of tokens generated per second.\footnote{We use 50 random prompts and let models generate up to 256 tokens, on an L40S GPU.} Our approach introduces only a 0.8\% drop in decoding speed (39.6 vs. 39.9 tokens/sec).

Still, since our compositional next-token prediction approach occurs in two stages---first deciding on likely candidates for \bases and \transformations---it naturally allows for optimizations like pruning base-form candidates before computing the logits over the full vocabulary~\citep{Holtzman2020topp}, which could further decrease runtime.

\begin{table}[t]
\centering
\scriptsize
\begin{tabular}{@{}lccccc@{}}
\toprule
Targets & Exact & Base & Diff.\ infl. & Other & $N$ \\
\midrule
Inflected targets & 81.47 & 15.77 & 1.96 & 0.80 & 9.4k \\
Base targets & 99.98 & -- & -- & 0.02 & 14k \\
\bottomrule
\end{tabular}
\caption{\patchscopes outcome rates for English targets. Errors are dominated by collapse to the base form, whereas confusion with a \emph{different} inflection of the same base is rare.}
\label{tab:appendix_patchscopes_errors}
\end{table}

\begin{table}[t]
\centering
\small
\begin{tabular}{lcccc}
\toprule
Space & Self sim. & Other sim. & Margin & Top-1 acc. \\
\midrule
Input & 0.186 & 0.040 & 0.146 & 98.21 \\
Output & 0.289 & 0.108 & 0.181 & 95.63 \\
\bottomrule
\end{tabular}
\caption{Analysis of the separability of initialized \transformation vectors across the eight English \transformation types, using individual inflection-base offsets (e.g., \textit{walked}-\textit{walk}) and their corresponding labels (e.g., \textsl{past tense}). Self-transformation cosine similarity exceeds cross-transformation similarity in both input and output spaces, with high top-1 classification accuracy.}
\label{tab:appendix_offset_geometry_summary}
\end{table}

\subsection{Compositional representation of morphology}
\label{appendix_inverse_scaling}

For regular embedding models (top panel in \figref{vocab_scaling_analysis}), we use \texttt{Llama2-7B}~\citep{llama2}, \texttt{Mistral-7B}~\citep{mistral7b}, \texttt{OLMo2-7B}~\citep{olmo2}, \texttt{Phi4-14B}~\citep{phi4}, \texttt{Llama3-8B}~\citep{llama3}, \texttt{Qwen2.5-7B}~\citep{qwen2.5}, and \texttt{Falcon3-7B}~\citep{falcon3}. For tied input-output embedding models (bottom panel), where input and output embeddings share parameters, we analyze \texttt{Llama3-3B}~\citep{llama3}, \texttt{Qwen2.5-3B}~\citep{qwen2.5}, \texttt{Phi4-Mini-Instruct-4B}~\citep{phi4}, \texttt{Aya-Expanse-8B}~\citep{ayaexpanse}, \texttt{Gemma2-2B} and \texttt{Gemma2-9B}~\citep{gemma2}.

\section{Experimental Details}
\label{appendix_experimental}

\subsection{Post-hoc Fine-tuning Details}
\label{appendix_implementation}
For fine-tuning, we use a learning rate of $5e-5$, a warmup ratio of $0.03$, a weight decay of $0.0$, and a sequence length of $m=256$. We train on $20k$ examples for 1 epoch. We run the post-hoc adaptation experiments on a single L40S GPU, with fine-tuning taking roughly 30 minutes, and inference taking up to 1 hour.

\subsection{Pretraining Details}
\label{appendix_pretraining}
This appendix provides implementation and evaluation details for the pretraining experiment in \secref{pretraining}.

\textbf{Implementation.} We follow the default hyperparameters in the \texttt{modded-nanoGPT} codebase~\cite{modded_nanogpt_2024}. All pretraining runs use 4 L40S GPUs, except for training on 1B tokens.

\textbf{Compositional tokenizer and coverage.} We start from a 50k-token GPT-2 tokenizer and remove any token that can be expressed as a base form plus \transformations, including a whitespace-prefix \transformation to capture pairs like `` walking'' vs.\ ``walking''. Unlike our post-hoc setup, we do not filter out compositions based on \patchscopes interpretation failures (\appref{appendix_filtering}).

\textbf{Factorized next-token prediction.} In the pretraining setting, each surface word is represented as a base form together with one choice from each transformation group, including a null label when no transformation from that group is active. If $w_i = (b_i, t_i^{(1)}, \ldots, t_i^{(G)})$, then the model factorizes
\begin{equation}
    p(w_i \mid w_{<i}) = p(b_i \mid w_{<i}) \prod_{g=1}^{G} p\!\left(t_i^{(g)} \mid b_i, w_{<i}\right)
\end{equation}
The model first computes base logits from the final hidden state $\mathbf{h}_i$,
\begin{equation}
    \begin{aligned}
        \boldsymbol{\ell}^{\text{base}}_i &= W_{\text{base}} \mathbf{h}_i + \mathbf{b}_{\text{base}}, \\
        p(b_i \mid w_{<i}) &= \mathrm{softmax}(\boldsymbol{\ell}^{\text{base}}_i)
    \end{aligned}
\end{equation}
It then predicts each transformation group conditioned on the selected base by passing $\mathbf{h}_i$ together with the chosen base's unembedding vector $\mathbf{u}_{b_i}$ to a transformation head, yielding group-wise logits
\begin{equation}
    \begin{aligned}
        \boldsymbol{\ell}^{(g)}_i &= f_g(\mathbf{h}_i, \mathbf{u}_{b_i}), \\
        p\!\left(t_i^{(g)} \mid b_i, w_{<i}\right) &= \mathrm{softmax}(\boldsymbol{\ell}^{(g)}_i)
    \end{aligned}
\end{equation}
During training, under teacher forcing, the transformation heads are conditioned on the gold base token. The negative log-likelihood therefore decomposes into a base-prediction term and a sum of transformation-group terms:
\begin{equation}
    \begin{aligned}
        \mathcal{L}
        &=
        -\sum_i \Bigg[
            \log p(b_i \mid w_{<i}) \\
        &
            +
            \sum_{g=1}^{G} \log p\!\left(t_i^{(g)} \mid b_i, w_{<i}\right)
        \Bigg]
    \end{aligned}
\end{equation}
At inference time, we first sample a base token, and then sample one transformation value from each group (for the experiments in this paper, both sampling operations use argmax), and finally compose them back into the surface realization. In other words, next-token prediction is hierarchical rather than a single softmax over surface forms.

\textbf{Bits-Per-Byte (BPB).} For both baseline and compositional models, we report BPB, computed as the average negative log-likelihood divided by the number of UTF-8 bytes in the evaluation text (lower is better). For the compositional model, we use the teacher-forced joint likelihood under the factorized distribution.

\begin{table*}
\centering
\resizebox{0.98\textwidth}{!}{%
\begin{tabular}{l@{\hspace{0.5em}}l@{\hspace{0.5em}}cc@{\hspace{0.5em}}cc@{\hspace{0.5em}}cc@{\hspace{0.5em}}cc@{\hspace{0.5em}}cc}
\toprule
& \multirow{2}{*}{{Language}} & \multicolumn{2}{c}{{Capitalization}} & \multicolumn{2}{c}{{Noun Inflection}} & \multicolumn{2}{c}{{Adjective Inflection}} & \multicolumn{2}{c}{{Verb Inflection}} & \multicolumn{2}{c}{{Derivation}} \\
\cmidrule(lr){3-4} \cmidrule(lr){5-6} \cmidrule(lr){7-8} \cmidrule(lr){9-10} \cmidrule(lr){11-12}
 & & \emph{In-Vocab.} & \emph{Out-Vocab.} & \emph{In-Vocab.} & \emph{Out-Vocab.} & \emph{In-Vocab.} & \emph{Out-Vocab.} & \emph{In-Vocab.} & \emph{Out-Vocab.} & \emph{In-Vocab.} & \emph{Out-Vocab.} \\
\midrule
\emph{ALLaM} & \hspace{0.5em}Arabic & --- & --- & 78\% (1.8k) & 16\% (3.6k) & 69\% (0.5k) & 25\% (1.0k) & 43\% (1.0k) & 15\% (2.7k) & --- & --- \\ \addlinespace[0.4em]
\emph{EuroLLM} & \hspace{0.5em}German & 100\% (0.2k) & 89\% (0.4k) & --- & --- & 27\% (0.3k) & 11\% (1.3k) & 88\% (0.3k) & 44\% (1.2k) & --- & --- \\
 & \hspace{0.5em}Russian & 98\% (66) & 96\% (0.7k) & 72\% (0.6k) & 28\% (4.2k) & 100\% (50) & 93\% (94) & 100\% (6) & 50\% (10) & --- & --- \\
 & \hspace{0.5em}Spanish & 100\% (1.0k) & 97\% (2.8k) & 83\% (0.7k) & 59\% (1.9k) & 82\% (0.5k) & 67\% (1.1k) & 72\% (0.8k) & 42\% (6.9k) & 46\% (65) & 20\% (0.4k) \\ \addlinespace[0.5em]
\emph{Llama-3} & \hspace{0.4em}English & 89\% (6.0k) & 85\% (8.4k) & 93\% (2.4k) & 63\% (5.6k) & 89\% (61) & 32\% (0.9k) & 85\% (0.9k) & 34\% (6.4k) & 34\% (41) & 4\% (12.8k) \\
\bottomrule
\end{tabular}
}
\caption{Accuracy of \patchscopes \emph{detokenization} interpretations for compositional input embeddings across languages.}
\label{tab:multilingual_patchscopes_detokenization}
\end{table*}

\section{Downstream Evaluation}
\label{appendix_eval}

We include 5 in-context examples for every task.
For each dataset, we use 5,000 examples (or the maximum available as some datasets have fewer available samples).
\paragraph{ARC} features 4-option multiple-choice science questions from grades 3 through 9. It has two subsets: ARC-Easy, focused on basic science knowledge, and ARC-Challenge, which involves more complex, procedural reasoning \citep{clark-etal-2018-think}.
\paragraph{BoolQ} comprises naturally occurring yes/no questions accompanied by passages that support the answer \citep{clark-etal-2019-boolq}.
\paragraph{COPA} offers binary multiple-choice questions centered around causal and consequential reasoning \citep{kavumba-etal-2019-choosing}.
\paragraph{HellaSwag} includes 4-option multiple-choice questions where the task is to select the most plausible continuation of a given context \citep{zellers-etal-2019-hellaswag}.
\paragraph{MMLU} presents 4-option multiple-choice questions across 57 subject areas, testing both factual knowledge and reasoning skills \citep{hendryckstest2021}.
\paragraph{PIQA} provides multiple-choice questions designed to evaluate physical commonsense understanding \citep{Bisk2020}.
\paragraph{SQuAD} pairs reading passages with related questions, where the correct answer is always a text span from the passage itself \citep{rajpurkar-etal-2016-squad}.
\paragraph{TriviaQA} features open-domain questions aimed at assessing general world knowledge \citep{joshi-etal-2017-triviaqa}.
\paragraph{Winogrande} contains questions modeled after the Winograd schema but scaled up in size and difficulty \citep{sakaguchi2021winogrande}.
\paragraph{XNLI} provides natural language inference examples in 15 languages, where the task is to determine whether a hypothesis is entailed by, contradicts, or is neutral with respect to a given premise \citep{conneau-etal-2018-xnli}.
\paragraph{XQuAD} is a cross-lingual question answering dataset that pairs reading passages with related questions in 11 languages, where the correct answer is always a text span from the passage itself \citep{artetxe-etal-2020-xquad}.
\paragraph{Global MMLU} extends the original MMLU benchmark to assess multilingual capabilities, featuring 4-option multiple-choice questions across 57 subject areas in 42 languages including low-resource languages, testing both factual knowledge and reasoning skills in diverse linguistic contexts \citep{singh-etal-2025-globalmmlu}.

\begin{table}[t!]
\centering
\resizebox{0.98\columnwidth}{!}{%
\begin{tabular}{l@{\hspace{0.5em}}ccc@{\hspace{0.5em}}ccc}
\toprule
\textbf{Transformation} & \multicolumn{3}{c}{In-vocab.} & \multicolumn{3}{c}{Out-of-vocab.} \\
\cmidrule(lr){2-4} \cmidrule(lr){5-7}
 & \emph{embed} & \emph{detok} & $N$ & \emph{embed} & \emph{detok} & $N$ \\
\midrule
\multicolumn{7}{l}{\textbf{Inflection}} \\
\hspace{1em}Plural (N) & 92\% & 92\% & 0.8k & 24\% & 31\% & 3.4k \\
\hspace{1em}\shortstack[l]{Plural (N) \\\hspace{0.5em} \& Present Singular (V)} & 86\% & 87\% & 1.6k & 35\% & 44\% & 2.1k \\
\hspace{1em}Present Singular (V) & 91\% & 91\% & 0.1k & 54\% & 64\% & 0.3k \\
\hspace{1em}Past (V) & 65\% & 68\% & 0.6k & 10\% & 15\% & 2.9k \\
\hspace{1em}Past Participle (V) & 79\% & 79\% & 14 & 24\% & 29\% & 21 \\
\hspace{1em}Gerund (V) & 83\% & 84\% & 0.2k & 17\% & 22\% & 3.2k \\
\hspace{1em}Superlative (ADJ) & 87\% & 87\% & 31 & 3\% & 10\% & 0.4k \\
\hspace{1em}Comparative (ADJ) & 47\% & 67\% & 30 & 4\% & 12\% & 0.4k \\
\midrule
\textbf{Capitalization} & 72\% & 73\% & 6.0k & 74\% & 76\% & 8.3k \\
\midrule
\multicolumn{7}{l}{\textbf{Derivation}} \\
\hspace{1em}-y & 17\% & 22\% & 18 & 2\% & 6\% & 1.5k \\
\hspace{1em}-er & 25\% & 25\% & 12 & 1\% & 3\% & 2.6k \\
\hspace{1em}-al & 62\% & 62\% & 8 & 1\% & 2\% & 0.7k \\
\hspace{1em}un- & 0\% & 33\% & 3 & 0\% & 1\% & 3.3k \\
\hspace{1em}re- & 67\% & 67\% & 3 & 0\% & 1\% & 1.8k \\
\hspace{1em}-ic & 100\% & 100\% & 2 & 5\% & 7\% & 0.4k \\
\hspace{1em}All derivatives & 40\% & 44\% & 52 & 0\% & 1\% & 31.4k \\
\bottomrule
\end{tabular}
}
\caption{Accuracy of \patchscopes interpretations for \texttt{Qwen-2.5-7B}.}
\label{tab:patchscopes_Qwen}
\end{table}

\begin{table}[t]
\centering
\resizebox{0.98\columnwidth}{!}{%
\begin{tabular}{l@{\hspace{0.5em}}ccc@{\hspace{0.5em}}ccc}
\toprule
\textbf{Transformation} & \multicolumn{3}{c}{In-vocab.} & \multicolumn{3}{c}{Out-of-vocab.} \\
\cmidrule(lr){2-4} \cmidrule(lr){5-7}
 & \emph{embed} & \emph{detok} & $N$ & \emph{embed} & \emph{detok} & $N$ \\
\midrule
\multicolumn{7}{l}{\textbf{Inflection}} \\
\hspace{1em}Plural (N) & 93\% & 94\% & 0.8k & 34\% & 42\% & 3.4k \\
\hspace{1em}\shortstack[l]{Plural (N) \\\hspace{0.5em} \& Present Singular (V)} & 86\% & 90\% & 1.6k & 41\% & 58\% & 2.1k \\
\hspace{1em}Present Singular (V) & 90\% & 91\% & 0.1k & 60\% & 71\% & 0.3k \\
\hspace{1em}Past (V) & 74\% & 85\% & 0.6k & 12\% & 24\% & 2.9k \\
\hspace{1em}Past Participle (V) & 100\% & 100\% & 14 & 24\% & 43\% & 21 \\
\hspace{1em}Gerund (V) & 93\% & 97\% & 0.2k & 26\% & 38\% & 3.2k \\
\hspace{1em}Superlative (ADJ) & 97\% & 97\% & 31 & 20\% & 38\% & 0.4k \\
\hspace{1em}Comparative (ADJ) & 87\% & 90\% & 30 & 7\% & 18\% & 0.4k \\
\midrule
\textbf{Capitalization} & 80\% & 96\% & 6.0k & 50\% & 85\% & 8.3k \\
\midrule
\multicolumn{7}{l}{\textbf{Derivation}} \\
\hspace{1em}-y & 65\% & 65\% & 17 & 13\% & 19\% & 1.5k \\
\hspace{1em}-er & 25\% & 33\% & 12 & 6\% & 19\% & 2.6k \\
\hspace{1em}-al & 75\% & 88\% & 8 & 4\% & 11\% & 0.7k \\
\hspace{1em}un- & 33\% & 33\% & 3 & 1\% & 6\% & 3.3k \\
\hspace{1em}re- & 100\% & 100\% & 3 & 1\% & 17\% & 1.8k \\
\hspace{1em}-ic & 100\% & 100\% & 2 & 10\% & 15\% & 0.4k \\
\hspace{1em}All derivatives & 63\% & 67\% & 51 & 2\% & 6\% & 31.4k \\
\bottomrule
\end{tabular}

}
\caption{Accuracy of \patchscopes interpretations for \texttt{OLMo-2-7B}.}
\label{tab:patchscopes_OLMo}
\end{table}

\begin{table}[t!]
\centering
\resizebox{0.98\columnwidth}{!}{%
\begin{tabular}{llcccc}
    \toprule
        \textbf{Category} & \textbf{Task} & \textbf{Baseline} & \textbf{End-to-end} & \textbf{$\Delta$} \\\midrule
       Knowledge & MMLU \textsubscript{(Acc.)} & 74.2 & 74.0 & -0.2 \\
        & ARC \textsubscript{(Acc.)} & 59.2 & 57.5 & -1.7 \\
        \midrule
       Reading Comprehension & BoolQ \textsubscript{(Acc.)} & 87.5 & 88.0 & +0.5 \\
        & TriviaQA \textsubscript{(EM)} & 58.3 & 56.1 & -2.2 \\
        & SQuAD \textsubscript{(EM)} & 37.3 & 36.2 & -1.1 \\
        \midrule
       Commonsense & Hellaswag \textsubscript{(Acc.)} & 59.6 & 58.4 & -1.2 \\
        & Winogrande \textsubscript{(Acc.)} & 75.5 & 75.0 & -0.5 \\
        & PIQA \textsubscript{(Acc.)} & 79.5 & 78.8 & -0.7 \\
        & COPA \textsubscript{(Acc.)} & 91.0 & 91.0 & +0.0 \\
        \midrule
         Average & & \textbf{69.1} & \textbf{68.3} & {-0.8} \\
    \bottomrule
    \end{tabular}
}
\caption{Downstream performance of English compositional-vocabulary models~(\textsl{End-to-end}) and their original, unmodified version~(\textsl{Baseline}) for \texttt{Qwen2.5-7B}.}
\label{tab:downstream_qwen}
\end{table}

\begin{table}[t!]
\centering
\resizebox{0.98\columnwidth}{!}{%
\begin{tabular}{llcccc}
    \toprule
        \textbf{Category} & \textbf{Task} & \textbf{Baseline} & \textbf{End-to-end} & \textbf{$\Delta$} \\\midrule
       Knowledge & MMLU \textsubscript{(Acc.)} & 62.7 & 62.2 & -0.5 \\
        & ARC \textsubscript{(Acc.)} & 60.5 & 58.4 & -2.1 \\
        \midrule
       Reading Comprehension & BoolQ \textsubscript{(Acc.)} & 84.4 & 84.4 & +0.0 \\
        & TriviaQA \textsubscript{(EM)} & 65.4 & 61.1 & -4.3 \\
        & SQuAD \textsubscript{(EM)} & 39.9 & 36.9 & -3.0 \\
        \midrule
       Commonsense & Hellaswag \textsubscript{(Acc.)} & 61.1 & 58.6 & -2.5 \\
        & Winogrande \textsubscript{(Acc.)} & 77.3 & 77.2 & -0.1 \\
        & PIQA \textsubscript{(Acc.)} & 80.2 & 79.2 & -1.0 \\
        & COPA \textsubscript{(Acc.)} & 90.0 & 91.0 & +1.0 \\
        \midrule
         Average & & \textbf{69.0} & \textbf{67.7} & {-2.3} \\
    \bottomrule
    \end{tabular}
}
\caption{Downstream performance for \texttt{OLMo-2-7B}.}
\label{tab:downstream_olmo}
\end{table}

\section{Additional Related Work}

\paragraph{Tokenization for morphologically-rich languages}
Standard BPE tokenization often struggles to capture morphologically complex languages~\citep{klein-tsarfaty-2020-getting,park-etal-2021-morphology,mager-etal-2022-bpe,hofmann-etal-2022-embarrassingly}. Arabic inflection, for instance, uses non-concatenative morphology that breaks standard subword reusability~\citep{Alyafeai2021EvaluatingVT,alrefaie2024exploring,tsarfaty-etal-2019-whats,gazit-etal-2025-splintering}.
Compositional vocabularies can bypass such limitations by representing surface forms as transformations over lexical roots, enabling reuse of base forms even when their surface realizations use diverging token sequences.

\end{document}